\documentclass[runningheads]{llncs}

\usepackage[T1]{fontenc}

\usepackage{booktabs}
\usepackage{multirow}
\usepackage{adjustbox}
\usepackage{graphicx,verbatim}
\usepackage{graphicx}
\usepackage{caption} 

\usepackage{amsmath}
\usepackage{amssymb}
\usepackage{booktabs}
\usepackage{array}
\usepackage{makecell}
\usepackage{color}
\usepackage{multirow}
\usepackage{tabularx}
\usepackage{hyperref}
\usepackage{url}
\usepackage[table,xcdraw]{xcolor}
\usepackage{amsmath}
\usepackage{xcolor}
\usepackage{graphicx} 
\usepackage{booktabs}
\usepackage[misc]{ifsym}

\begin{document}

\title{ClinRAG-GRAPH: Clinical-prior Retrieval-Augmented Graph Model with Domain Adversarial Learning for Breast pCR Prediction}

\author{Yaofei Duan\inst{1,2,3} \and
Yuhao Huang \inst{4,5} \and
Tianyu Zhang \inst{1,2} \and
Yuan Gao \inst{2,6} \and
Luyi Han \inst{1,2} \and 
Xin Wang \inst{2,6} \and
Xinyu Xie \inst{3} \and
Xinglong Liang \inst{1,2} \and
Chunyao Lu \inst{1,2} \and
Muzhen He \inst{7} \and
Patrick Pang \inst{3} \and 
Yue Sun \inst{3} \and
Ning Mao \inst{8} \and
Tao Tan \inst{3}\textsuperscript{(\Letter)} \and 
Ritse Mann \inst{1,2}
}
\authorrunning{Y. Duan et al.}

\institute{Department of Medical Imaging, Radboud University Medical Center,  Nijmegen, The Netherlands \and
The Netherlands Cancer Institute, Amsterdam, The Netherlands \and
Faculty of Applied Sciences, Macao Polytechnic University, Macau, China\\
\email{taotan@mpu.edu.mo}\\ \and
Boston Children's Hospital, Harvard Medical School, Boston, USA \\
\and
Centre for Artificial Intelligence and Robotics, Hong Kong Institute of Science \& Innovation, Chinese Academy of Sciences, Hong Kong, China 
\and
Imaging Division, University Medical Center Utrecht, Utrecht, The Netherlands \and
Department of Radiology, Fuzhou University Affiliated Provincial Hospital, Fuzhou, China \and
Department of Radiology, Yantai Yuhuangding Hospital, Shandong, China 
}

\titlerunning{ClinRAG-GRAPH}

\maketitle              
\begin{abstract}
Neoadjuvant chemotherapy (NAC) response prediction is clinically important for treatment stratification in breast cancer.
However, robust pre-treatment pathological complete response (pCR) prediction remains challenging due to insufficient cross-modal modeling, multicenter imaging heterogeneity, and weak evidence-grounded interpretability. 
We propose \textbf{ClinRAG-GRAPH}, a \textbf{Clin}ically informed \textbf{R}etrieval-\textbf{A}ugmented \textbf{G}eneration \textbf{Graph} framework, for pre-treatment pCR prediction from DCE-MRI, structured clinical variables, and biopsy-derived pathological biomarkers. 
ClinRAG-GRAPH constructs an intra-patient clinical-prior graph and applies a prior-guided relation-aware graph convolutional network for structured multimodal representation learning. 
To improve cross-center robustness, we introduce a dual-branch domain-adversarial learning strategy to suppress protocol-related MRI bias while preserving pCR-relevant features. 
To enhance interpretability, we further incorporate large language model (LLM)-driven subgraph RAG module that retrieves clinically analogous historical cases and integrates retrieved evidence for pCR inference. 
We assemble a large-scale multicenter NAC breast cancer cohort for extensive validation, drawing from two public sources and three in-house centers.
Results show that ClinRAG-GRAPH achieves AUCs of 0.815 on the internal test set and 0.774/0.712 on two external test sets, demonstrating robust pre-treatment pCR prediction across centers. 
The code is available at the anonymized link \url{https://github.com/miccai26-1181/ClinRAG-GRAPH}.


\end{abstract}
\section{Introduction}
Neoadjuvant chemotherapy (NAC) is a standard treatment for patients with locally advanced or high-risk breast cancer, and pathological complete response (pCR) after NAC is a clinically important endpoint associated with favorable prognosis in several breast cancer subtypes~\cite{spring2020pathologic,LMF}. 
However, pCR is determined only after surgery by pathological assessment. 
Therefore, accurate prediction of pCR from multimodal data may support individualized NAC treatment~\cite{krasniqi2025multimodal}.

Recent work on NAC response prediction has evolved from imaging-only models to multimodal frameworks that combine MRI with clinical and pathological information~\cite{nishizawa2025attention,xu2025deep}. 
MRI-based studies demonstrated the feasibility of imaging-driven prediction, while multicenter studies also revealed sensitivity to scanner and protocol heterogeneity, highlighting the risk of center-specific bias and limited cross-center generalization~\cite{joo2021multimodal,zhang2021prediction}. 
For pre-treatment pCR prediction, prior work has shown that integrating baseline MRI with clinical variables improves performance compared with unimodal settings~\cite{joo2021multimodal}. Subsequent studies further suggested that adding clinical and pathological information can improve predictive accuracy and reinforce the value of multimodal baseline integration~\cite{pesapane2021radiomics,lv2025deep}. These results support multimodal modeling as a promising direction for pCR prediction. 
While some recent methods consider longitudinal or multi-time data for pCR prediction~\cite{gao2024explainable,LMF,m2fusion}, we focus on the pre-treatment setting because it is most relevant for early treatment escalation for likely non-responders and de-escalation strategies for patients with high response probability.

However, several limitations remain for robust multicenter pre-treatment pCR prediction. 
First, many methods still rely on simple fusion operators (e.g., direct concatenation), which may not fully exploit clinically structured cross-modal dependencies. Related limitations have also been noted in recent multimodal studies~\cite{LMF,m2fusion}. 
Second, explicit domain-invariant learning to mitigate multicenter MRI heterogeneity is not consistently incorporated~\cite{zhang2021prediction}. 
Third, most pCR models output only a scalar pCR probability without case-based evidence retrieval, which limits evidence-grounded interpretation in clinical workflows, motivating retrieval-augmented generation (RAG)
inference strategies~\cite{amugongo2025retrieval}.

To address these limitations, we propose \textbf{ClinRAG-GRAPH}, a clinically informed retrieval-augmented graph framework for pre-treatment pCR prediction from multimodal data, including DCE-MRI, structured clinical variables, and biopsy-derived pathological biomarkers. 
The main contributions of our work are as follows. 
First, we encode clinically motivated prior knowledge into the graph construction and learn multimodal features with a relation-aware attention graph convolutional network (GCN). 
Second, we design adversarial learning strategy to mitigate multicenter MRI protocol heterogeneity and improve task-relevant representations. We further incorporate an LLM-based subgraph RAG module to support evidence-grounded and interpretable pCR inference. 
Third, we conduct a multicenter evaluation on a large-scale NAC breast cancer cohort comprising two public datasets and three in-house datasets, demonstrating competitive and consistent performance against five representative baselines.

\begin{figure}[t]
\centering
\includegraphics[width=1\textwidth]{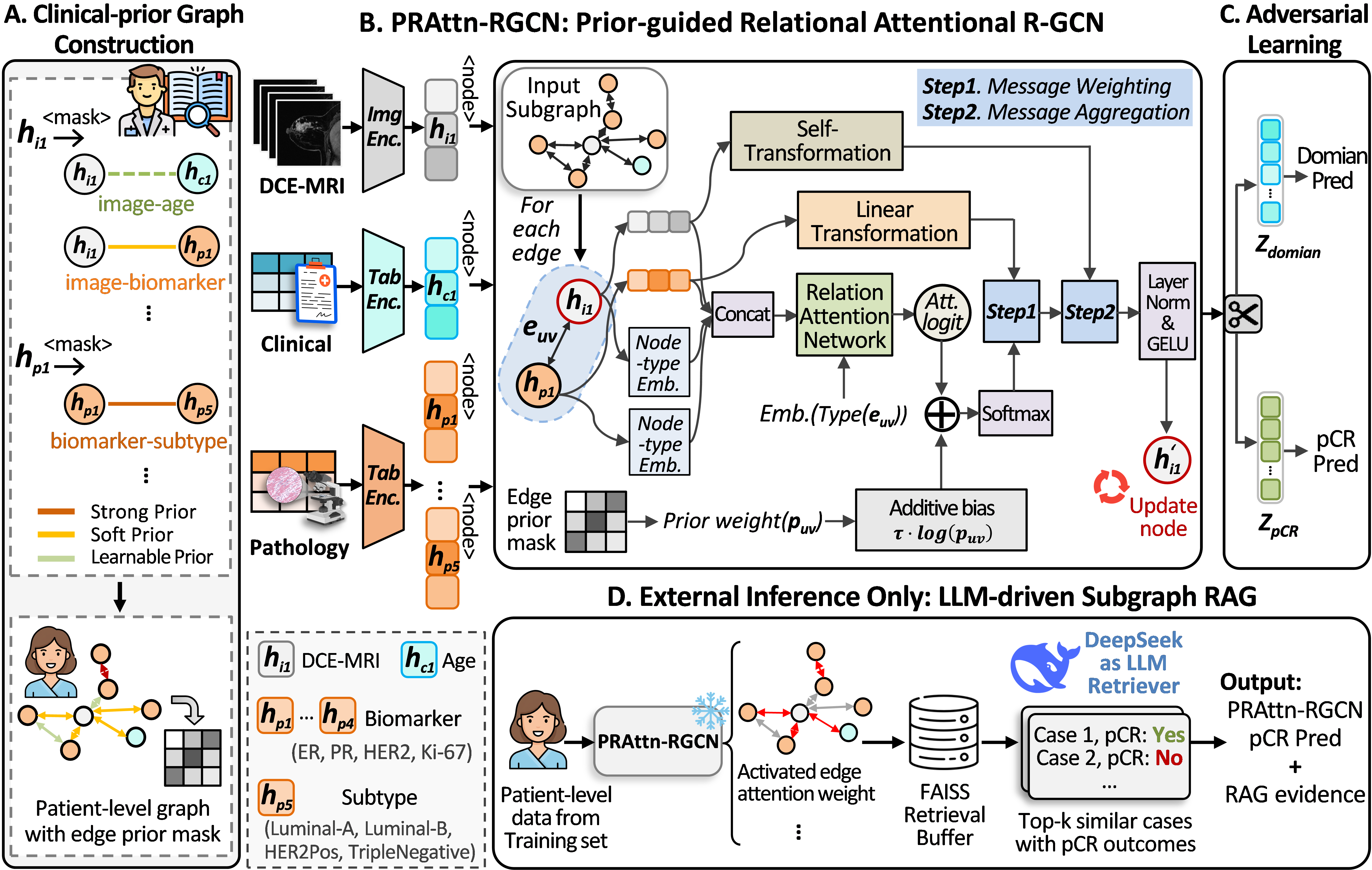}
\caption{Overview of ClinRAG-GRAPH for breast pCR prediction. Emb., Tab., and Enc. denote embedding, table, and encoder; \texttt{\textless mask/\textgreater} denotes the edge-prior mask, and scissors indicate gradient-reversal adversarial decoupling.}
\label{fig1}
\end{figure}

\section{Method} 
We proposed ClinRAG-GRAPH for multicenter breast pCR prediction by integrating heterogeneous multimodal data into a unified intra-patient graph. 
For each patient $i$, we denote the MRI volume as $X_i$, the clinical/pathological attributes as $t_i$, the binary pCR label as $y_i\in\{0,1\}$, and the acquisition center as $c_i\in\{1,\ldots,C\}$. Our goal is to learn a predictor that is discriminative for $y_i$ while being robust to domain shifts induced by $c_i$.

As shown in Fig. \ref{fig1}, ClinRAG-GRAPH first encodes the MRI volume using a DenseNet-121 encoder and constructs a clinical-prior graph $G_i$ with edge-wise prior strengths $p_{uv}$. It then applies a prior-guided relational attentional R-GCN to obtain a patient embedding and predict pCR, while enforcing center-invariant imaging representations via a gradient-reversal domain classifier. We further employ an LLM-driven case-based subgraph RAG module to retrieve clinically analogous historical cases based on learned embeddings and attention signatures, and then fuse the neighbor-derived estimate with the R-GCN prediction for a more evidence-grounded and robust inference.

\subsection{Hierarchical Semantic Clinical-prior Graph Construction}
In multi-modal pCR prediction, purely data-driven graph learning faces two practical challenges: (i) homogeneous connectivity (e.g., fully connecting image-table/table-table pairs) blurs heterogeneous multi-modal medical semantics, and (ii) enforcing priors via auxiliary losses constraints can destabilize optimization and introduce extra hyperparameters.

We therefore construct a hierarchical semantic clinical-prior graph at the patient level. For each patient $i$, we define an intra-patient directed graph $G_i=(V,E)$ with one imaging node and multiple clinical-$/$pathological- variable nodes. Each directed edge $(u,v)\in E$ is associated with a relation type $r(u,v)$ and a scalar prior weight $p_{uv}>0$ that explicitly encodes the confidence of medical knowledge about this interaction. Guided by clinical practice guidelines and radiologist consensus~\cite{loibl2024early,amin2017ajcc,gradishar2022breast,hurvitz2018neoadjuvant}, we organize edges into three semantic tiers:
\begin{itemize}
\item \textbf{Strong prior.} Guideline-level, stable associations (e.g., strong couplings
among key molecular biomarkers), encoded with a fixed large $p_{uv}=2$.
\item \textbf{Soft prior.} Clinically agreed relations with inter-patient variability, encoded with a relatively large initialization of $p_{uv}=0.8$.
\item \textbf{Learnable prior.} Uncertain relations, weakly initialized (small $p_{uv}=0.2$) and primarily determined by data (e.g., patient age with breast caner subtypes).
\end{itemize}

To inject priors without auxiliary loss terms, we incorporate $p_{uv}$ as a log-bias on attention logits only at the first graph layer (Sec.~2.2), providing early calibration toward clinically plausible message passing while keeping deeper layers fully learnable. We fix the temperature $t=0.2$ in all experiments.

\subsection{PRAttn-RGCN: Prior-guided Relational Attentional R-GCN}
Follow R-GCN~\cite{schlichtkrull2018modeling}, given $G_i=(V,E)$, we perform relation-aware attentive message passing to fuse the imaging node with heterogeneous clinical/pathological variable nodes. Each directed edge $(u\!\to\!v)$ carries a semantic relation type $r(u,v)$, and nodes belong to different modalities. The prior strength $p_{uv}$ is injected only at the first layer to guide early-stage cross-modal information flow.\\
\textbf{Relation-aware attention.}
Let $\mathbf{h}^{(l)}_v\in\mathbb{R}^d$ be the node representation at layer $l$, and $\mathbf{t}_{\tau(v)}$ be the embedding of node type $\tau(v)$. For each edge $(u\!\to\!v)$, we compute a relation-specific attention logit
\begin{equation}
e^{(l)}_{uv}=
\mathbf{a}_{r(u,v)}
(
[\mathbf{h}^{(l)}_{u};\mathbf{h}^{(l)}_{v};\mathbf{t}_{\tau(u)};\mathbf{t}_{\tau(v)}]
),
\label{eq:rel_attn_logit}
\end{equation}
where $\mathbf{a}_{r}(\cdot)$ is a lightweight relation-specific scorer. We inject the prior log-bias only when $l=0$:
\begin{equation}
\tilde e^{(l)}_{uv}=e^{(l)}_{uv}+\mathbb{I}[l=0]\cdot \tau \log(p_{uv}),
\qquad
\alpha^{(l)}_{uv}=\operatorname{softmax}_{u\in\mathcal{N}(v)}(\tilde e^{(l)}_{uv}).
\label{eq:rel_attn_weight}
\end{equation}
\\
\textbf{Relational message passing.}
Messages are transformed per relation and aggregated with attention:
\begin{equation}
\mathbf{h}^{(l+1)}_{v}=
\phi\!(
\mathbf{W}_{0}\mathbf{h}^{(l)}_{v}+
\sum_{u\in\mathcal{N}(v)}
\alpha^{(l)}_{uv}\,\mathbf{W}_{r(u,v)}\mathbf{h}^{(l)}_{u}
),
\label{eq:rel_rgcn_update}
\end{equation}
where $\mathbf{W}_{0}$ means self-loop transform and $\mathbf{W}_{r}$ is relation-specific (optionally using basis decomposition with large $|\mathcal{R}|$). $\phi(\cdot)$ denotes LayerNorm$\&$GELU. \\
\textbf{Graph readout.}
After $L$ layers, node embeddings are pooled to obtain a patient representation
$\mathbf{g}_i=\mathrm{MeanPool}(\{\mathbf{h}^{(L)}_v\}_{v\in V})$, which is fed to an Multilayer Perceptron (MLP) to predict pCR. The first-layer attention $\{\alpha^{(0)}_{uv}\}$ is retained as the attention signature for downstream case retrieval.

\subsection{Adversarial Learning for Decoupling MRI Domain Noise}
Multicenter DCE-MRI inevitably contains scanner/protocol-specific artifacts that can dominate image embeddings and harm cross-center generalization. To suppress such domain cues while preserving pCR-discriminative information, we impose a domain-adversarial constraint on the imaging representation. Let $\mathbf{h}_{\text{img}}^{(0)} = f_{\text{img}}(X_i) \in \mathbb{R}^d$ denotes the embedding used to initialize the imaging node. We attach an adversarial branch consisting of a lightweight projector $f_{\mathrm{adv}}(\cdot)$ and a domain classifier $D(\cdot)$, trained through a gradient reversal layer (GRL)~\cite{ganin2016domain}:
\begin{equation}
\mathbf{z}_{\text{adv}} = f_{\text{adv}}(\mathbf{h}_{\text{img}}^{(0)}), 
\qquad
\hat{\mathbf{c}} = D(\operatorname{GRL}(\mathbf{z}_{\text{adv}})),
\label{eq:adv_branch}
\end{equation}
where $\hat{\mathbf{c}}\in\mathbb{R}^{C}$ denotes the center logits and GRL multiplies the back-propagated gradient by a negative constant, enabling end-to-end optimization.
Then, the training objective combines pCR supervision and domain confusion is:
\begin{equation}
\mathcal{L}_{\mathrm{total}}
=\mathcal{L}_{\mathrm{pCR}}(\hat{y}_i, y_i)
+\lambda_{\mathrm{adv}} \mathcal{L}_{\mathrm{adv}}(\hat{\mathbf{c}}, c_i),
\end{equation}
where $\mathcal{L}_{\mathrm{pCR}}$ is BCE loss, $\mathcal{L}_{\mathrm{adv}}$ is CE loss with $\lambda_{\mathrm{adv}}$ set to 0.05 through empirical tuning.

\subsection{LLM-driven Subgraph RAG for Evidence-grounded Inference}
To improve the robustness under distribution shift, we augment the trained PRAttn-RGCN with an LLM (\textbf{DeepSeek-V3.2}~\cite{deepseek}) retriever, design an RAG module that grounds inference on similar patient evidence. After training, we infer the training set and export intermediate results as an evidence buffer. For each patient $i$, we store (i) a graph-level embedding $\mathbf{g}_i$, (ii) an MRI embedding $\mathbf{z}_i$, (iii) a clinical-pathological embedding $\mathbf{m}_i$, and (iv) edge-level attention weights $\boldsymbol{\alpha}_i$ from the frozen PRAttn-RGCN. We then build FAISS~\cite{johnson2019billion} inner-product indices over $\ell_2$-normalized embeddings for approximate nearest-neighbor search.

At inference, a query patient $q$ is encoded as $(\mathbf{g}_q,\mathbf{z}_q,\mathbf{m}_q,\boldsymbol{\alpha}_q)$. DeepSeek outputs a schema-constrained retrieval plan from query-side signals (e.g., candidate pool sizes, filters (reduce false neighbors), fusion weights, and top-$K$), without making predictions. Candidates are retrieved via multi-route search and ranked by a fused score combining representation similarity and a subgraph match induced by $\boldsymbol{\alpha}$. The top-$K$ neighbors are deterministically aggregated and combined with the PRAttn-RGCN output to yield the final prediction, enabling evidence-grounded inference while preserving backbone calibration. 
Optionally, DeepSeek generates a structured post-hoc explanation based solely on retrieval metadata, fused scores, and driver-edge cues, exported as a verifiable JSON record.

\section{Experiments}
\textbf{Datasets and Implementation Details.}
We evaluated ClinRAG-GRAPH on two public datasets and three in-house datasets, with dataset statistics summarized in Table~\ref{dataset}. 
Specifically, data from three source centers (DUKU~\cite{duke}, ISPY1~\cite{ispy1}, and Zcenter) were used for model development and split at the patient level in a 7:1:2 ratio for training, validation, and internal testing, respectively, while Qcenter and Ycenter were held out for external generalization evaluation. 
All models were implemented in \textit{PyTorch} and trained on an NVIDIA A40 GPU (48 GB memory). 
For preprocessing, multi-phase DCE-MRI scans were cropped to tumor-centered regions and concatenated along the temporal dimension to form an input volume of size \((192, 256, 256)\), with intensities processed using z-score normalization. 
The network was optimized using AdamW (learning rate \(2\times10^{-5}\), weight decay \(1\times10^{-4}\), batch size 12) for 50 epochs.

\textbf{Evaluation Metrics}.
The quantitative evaluation was performed using standard classification metrics, i.e., AUC (95\% CI) to assess discrimination, balanced accuracy (bAcc) to account for class imbalance, sensitivity (SEN) to evaluate positive-case detection, and specificity (SPE) to assess negative-case recognition.

\begin{table}[t]
\centering
\caption{Multicenter dataset summary across \textcolor{red}{internal} and \textcolor{blue}{external} domains.}
\label{dataset}
\renewcommand{\arraystretch}{1.}
\setlength{\tabcolsep}{6pt}
\small
\begin{adjustbox}{max width=0.7\textwidth}
\begin{tabular}{l|l|ccccc}
\toprule
\multicolumn{2}{l|}{Domain} & \textcolor{red}{DUKE}~\cite{duke} & \textcolor{red}{ISPY1}~\cite{ispy1} & \textcolor{red}{Zcenter} & 
\textcolor{blue}{Ycenter} & \textcolor{blue}{Qcenter} \\
\midrule
\multirow{2}{*}{Amount} 
& pCR     & 62  & 49  & 83  & 54  & 33 \\
& non-pCR & 235 & 123 & 367 & 166 & 88 \\
\midrule
\multicolumn{2}{l|}{Public available} & Yes & Yes & No & No & No \\
\bottomrule
\end{tabular}
\end{adjustbox}
\end{table}

\begin{table}[t]
\centering
\caption{Performance comparison on internal and external test sets. AUC differences were evaluated using the paired DeLong test with a significance threshold of \textit{p<0.05}. Differences were significant at \textit{p<0.05} for all comparisons.}
\label{tab:main_results_compact}
\renewcommand{\arraystretch}{1.3}
\setlength{\tabcolsep}{0.4pt}
\small
\begin{adjustbox}{max width=\textwidth, center}
\begin{tabular}{lcccc cccc cccc}
\toprule
\textbf{Method}
& \multicolumn{4}{c}{\textbf{Internal (include 3 test sets)}}
& \multicolumn{4}{c}{\textbf{External Ycenter}}
& \multicolumn{4}{c}{\textbf{External Qcenter}} \\
\cmidrule(lr){2-5}\cmidrule(lr){6-9}\cmidrule(lr){10-13}
& \makecell[c]{\textbf{AUC \textcolor{red!85!black}{↑}} \\ (95\% CI)}
& \textbf{bAcc\textcolor{red!85!black}{↑}}
& \textbf{SEN}
& \textbf{SPE}
& \makecell[c]{\textbf{AUC \textcolor{red!85!black}{↑}} \\ (95\% CI)}
& \textbf{bAcc\textcolor{red!85!black}{↑}}
& \textbf{SEN}
& \textbf{SPE}
& \makecell[c]{\textbf{AUC \textcolor{red!85!black}{↑}} \\ (95\% CI)}
& \textbf{bAcc\textcolor{red!85!black}{↑}}
& \textbf{SEN}
& \textbf{SPE} \\
\midrule

R-GCN
& \makecell[c]{0.706 \\ (0.611-0.792)} & 0.668 & 0.500 & 0.837
& \makecell[c]{0.648 \\ (0.560-0.730)} & 0.575 & \textbf{0.241} & 0.910
& \makecell[c]{0.680 \\ (0.580-0.778)} & 0.530 & 0.242 & 0.818 \\

CLST
& \makecell[c]{0.539 \\ (0.450-0.626)} & 0.498 & 0.056 & \textbf{0.940}
& \makecell[c]{0.533 \\ (0.412-0.651)} & 0.491 & 0.061 & 0.921
& \makecell[c]{0.513 \\ (0.409-0.614)} & 0.532 & 0.214 & 0.850 \\

M2Fusion
& \makecell[c]{0.505 \\ (0.399-0.613)} & 0.534 & 0.167 & 0.902
& \makecell[c]{0.484 \\ (0.396-0.573)} & 0.455 & 0.037 & 0.873
& \makecell[c]{0.454 \\ (0.337-0.572)} & 0.491 & 0.061 & 0.921 \\

LMF
& \makecell[c]{0.678 \\ (0.584-0.768)} & 0.625 & 0.452 & 0.797
& \makecell[c]{0.693 \\ (0.614-0.767)} & 0.510 & 0.074 & 0.946
& \makecell[c]{0.666 \\ (0.565-0.762)} & 0.532 & 0.212 & 0.852 \\

iMRhpc
& \makecell[c]{0.699 \\ (0.605-0.790)} & 0.635 & 0.381 & 0.889
& \makecell[c]{0.682 \\ (0.598-0.761)} & 0.538 & 0.185 & 0.892
& \makecell[c]{0.703 \\ (0.605-0.792)} & 0.492 & 0.030 & \textbf{0.955} \\

\textbf{Ours}
& \makecell[c]{\textbf{0.815} \\ \textbf{(0.738-0.885)}} & \textbf{0.691} & \textbf{0.500} & 0.882
& \makecell[c]{\textbf{0.774} \\ \textbf{(0.700-0.843)}} & \textbf{0.575} & 0.204 & \textbf{0.946}
& \makecell[c]{\textbf{0.712} \\ \textbf{(0.604-0.815)}} & \textbf{0.676} & \textbf{0.455} & 0.898 \\
\bottomrule
\end{tabular}
\end{adjustbox}
\end{table}

\begin{table}[t]
\centering
\caption{Ablation study results for model components, subgraph RAG is applied only to external sets. CP: clinical prior; PR: PRAttn-RGCN; ADV: adversarial decoupling; RAG: DeepSeek-based subgraph RAG. AUC differences were evaluated using the paired DeLong test (\textit{p<0.05}). Differences were significant at \textit{p<0.05} except: V1 on Ycenter (\textit{p=0.074}), V2 on internal sets (\textit{p=0.062}).}
\label{tab:ablation_compact}
\renewcommand{\arraystretch}{1.3}
\setlength{\tabcolsep}{0.5pt}
\small 
\begin{adjustbox}{width=\textwidth, center}
\begin{tabular}{cccc cccc cccc cccc}
\toprule
\multicolumn{4}{c}{\textbf{Module}} 
& \multicolumn{4}{c}{\textbf{Internal (include 3 test sets)}}
& \multicolumn{4}{c}{\textbf{External Ycenter}} 
& \multicolumn{4}{c}{\textbf{External Qcenter}} \\
\cmidrule(lr){1-4}\cmidrule(lr){5-8}\cmidrule(lr){9-12}\cmidrule(lr){13-16}
\textbf{CP} & \textbf{PR} & \textbf{ADV} & \textbf{RAG}
& \makecell[c]{\textbf{AUC \textcolor{red!85!black}{↑}} \\ (95\% CI)} & \textbf{bAcc\textcolor{red!85!black}{↑}} & \textbf{SEN} & \textbf{SPE}
& \makecell[c]{\textbf{AUC \textcolor{red!85!black}{↑}} \\ (95\% CI)} & \textbf{bAcc\textcolor{red!85!black}{↑}} & \textbf{SEN} & \textbf{SPE}
& \makecell[c]{\textbf{AUC \textcolor{red!85!black}{↑}} \\ (95\% CI)} & \textbf{bAcc\textcolor{red!85!black}{↑}} & \textbf{SEN} & \textbf{SPE} \\
\midrule

\textbf{-} & \textbf{-} & \textbf{-} & \textbf{-}
& \makecell[c]{0.706 \\ (0.611-0.792)} & 0.668 & 0.500 & 0.837
& \makecell[c]{0.648 \\ (0.560-0.730)} & 0.575 & 0.241 & 0.910
& \makecell[c]{0.680 \\ (0.580-0.778)} & 0.530 & 0.242 & 0.818 \\

\checkmark & \textbf{-} & \textbf{-} & \textbf{-}
& \makecell[c]{0.745 \\ (0.653-0.830)} & 0.679 & \textbf{0.548} & 0.811
& \makecell[c]{0.708 \\ (0.628-0.784)} & \textbf{0.606} & 0.278 & 0.934
& \makecell[c]{0.683 \\ (0.583-0.780)} & 0.610 & 0.515 & 0.705 \\

\checkmark & \checkmark & \textbf{-} & \textbf{-}
& \makecell[c]{0.787 \\ (0.708-0.861)} & 0.645 & 0.452 & 0.837
& \makecell[c]{0.746 \\ (0.669-0.816)} & 0.603 & 0.296 & 0.910
& \makecell[c]{0.709 \\ (0.603-0.806)} & \textbf{0.697} & \textbf{0.576} & 0.818 \\

\checkmark & \checkmark & \checkmark & \textbf{-}
& \makecell[c]{\textbf{0.815} \\ \textbf{(0.738-0.885)}} & \textbf{0.691} & 0.500 & \textbf{0.882}
& \makecell[c]{0.768 \\ (0.695-0.835)} & 0.600 & \textbf{0.315} & 0.886
& \makecell[c]{0.703 \\ (0.593-0.804)} & 0.642 & 0.455 & 0.830 \\

\checkmark & \checkmark & \checkmark & \checkmark
& \textbf{--} & \textbf{--} & \textbf{--} & \textbf{--}
& \makecell[c]{\textbf{0.774} \\ \textbf{(0.700-0.843)}} & 0.575 & 0.204 & \textbf{0.946}
& \makecell[c]{\textbf{0.712} \\ \textbf{(0.604-0.815)}} & 0.676 & 0.455 & \textbf{0.898} \\

\bottomrule
\end{tabular}
\end{adjustbox}
\end{table}

\begin{figure}[t]
\centering
\begin{minipage}[t]{0.27\linewidth}
    \centering
    \includegraphics[width=\linewidth]{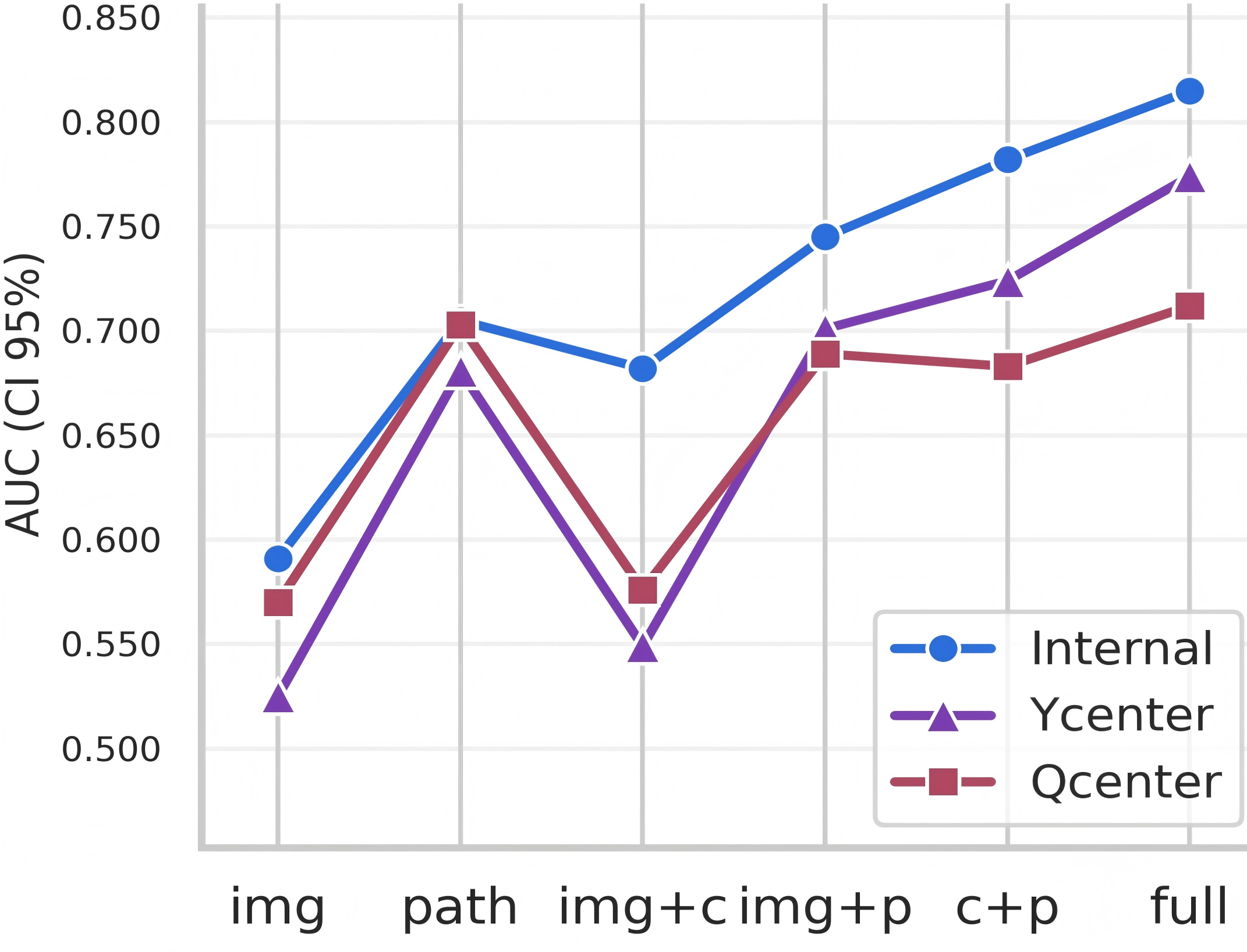}
    \caption{AUC results for modality pairs.}
    \label{fig:left}
\end{minipage}
\hfill
\begin{minipage}[t]{0.71\linewidth}
    \centering
    \includegraphics[width=\linewidth]{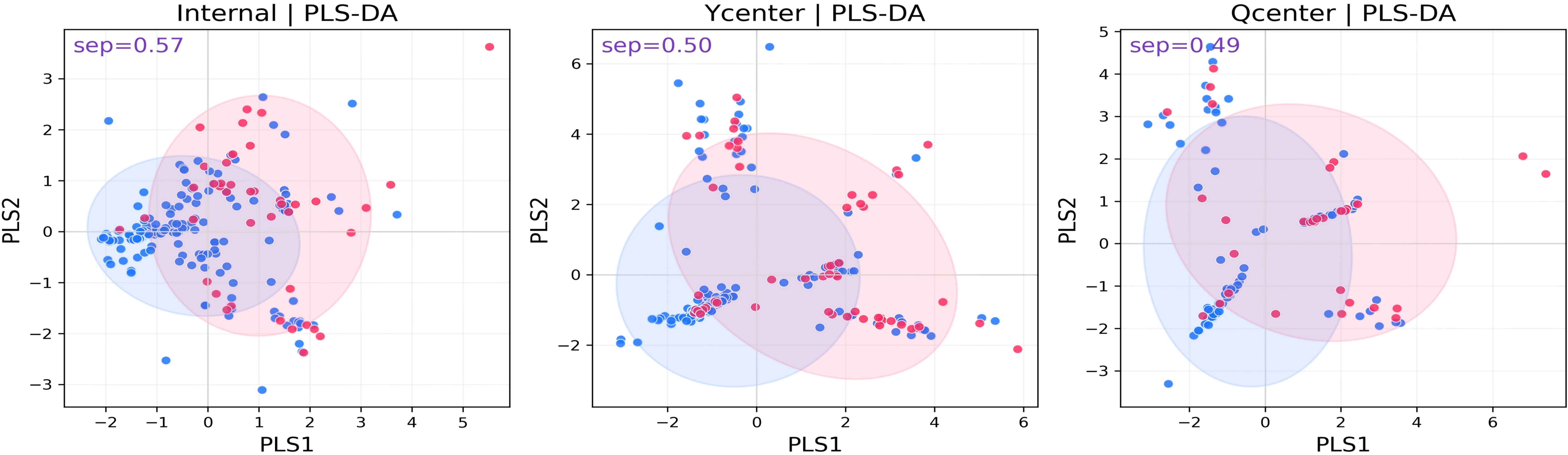}
    \caption{PLS-DA visualizations of binary class separation across internal and external cohorts.}
    \label{fig:right}
\end{minipage}
\end{figure}

\textbf{Internal Comparison.}
We evaluate ClinRAG-GRAPH against state-of-the-art methods including CLSTM~\cite{zhang2021prediction}, M2Fusion~\cite{m2fusion}, LMF~\cite{LMF}, iMRhpc~\cite{gao2024explainable}. 
As shown in Table \ref{tab:main_results_compact}, on the internal test set (aggregated from three internal test sets), ClinRAG-GRAPH achieves the best AUC of 0.815 (95\% CI: 0.738–0.885), outperforming R-GCN (0.706) and other fusion strategies such as LMF (0.678) and iMRhpc (0.699). It also improves balanced accuracy (0.691) and specificity (0.882) while maintaining sensitivity (0.500), indicating a more favorable trade-off. Paired DeLong tests confirm significant AUC gains ($p<0.05$).

\textbf{External evaluation.}
ClinRAG-GRAPH remains robust under cross-domain shift. On Ycenter, it attains the highest specificity (0.946) while preserving the best AUC, reducing false positives in unseen domains. On the more challenging Qcenter cohort, it yields the best balanced accuracy (0.676) and markedly higher sensitivity (0.455) than all baselines, reflecting improved detection without sacrificing overall discrimination. 
The PLS-DA projection in Fig.~\ref{fig:right} further shows that model effectively distinguishes pCR patients across internal and external cohorts. 
Overall, ClinRAG-GRAPH consistently enhances both in-domain performance and multicenter generalization for pre-treatment pCR prediction.

\textbf{Ablation Study.}
Table \ref{tab:ablation_compact} shows that each proposed component contributes to performance. Introducing the clinical-prior graph improves over the R-GCN backbone, highlighting the value of explicit clinical-pathological relational priors for multimodal fusion. PRAttn-RGCN and adversarial decoupling provide further gains, suggesting the benefits of prior-aware message passing and more domain-invariant MRI representations. In addition, LLM-driven subgraph RAG further improves performance on external test sets, indicating enhanced robustness under distribution shift. Modality ablation (Fig. \ref{fig:left}) further shows that pathology contributes most to pCR prediction.

\begin{figure}[t]
\centering
\includegraphics[width=1\textwidth]{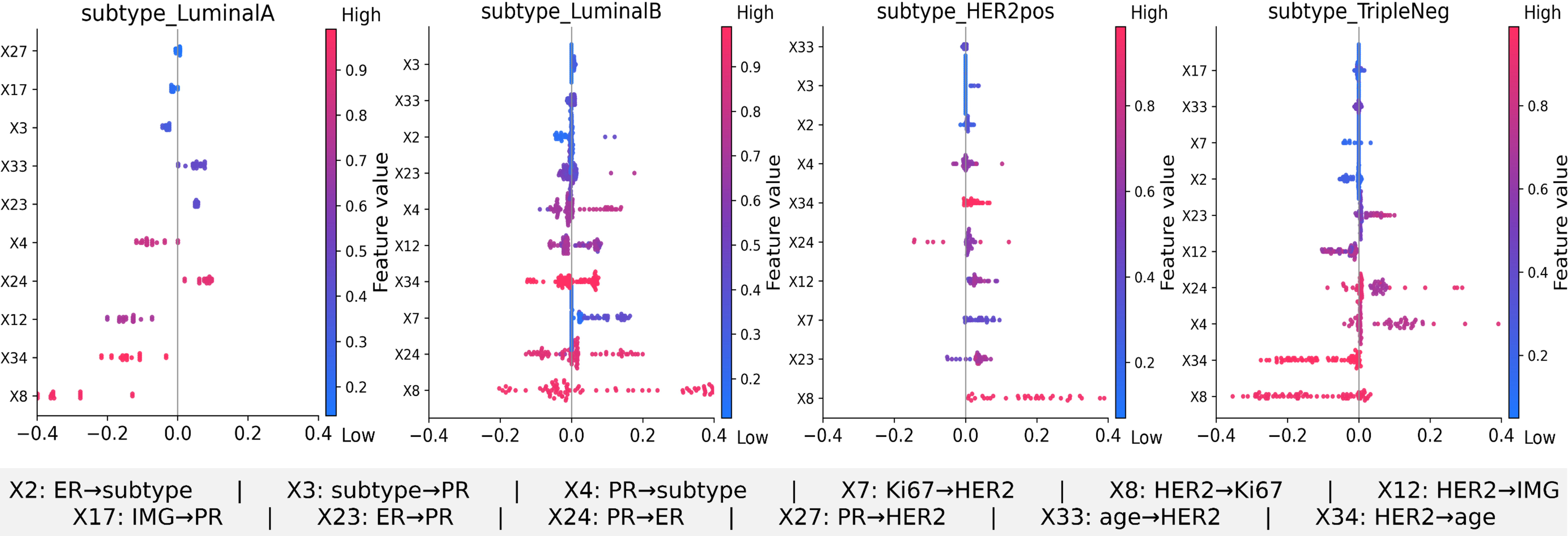}
\caption{SHAP analysis validating the contribution of key directed graph edges in the ClinRAG-GRAPH for LuminalA/B, HER2-pos, and TripleNeg subgroups.
} 
\label{SHAP}
\end{figure}

\begin{figure}[t]
\centering
\includegraphics[width=1\textwidth]{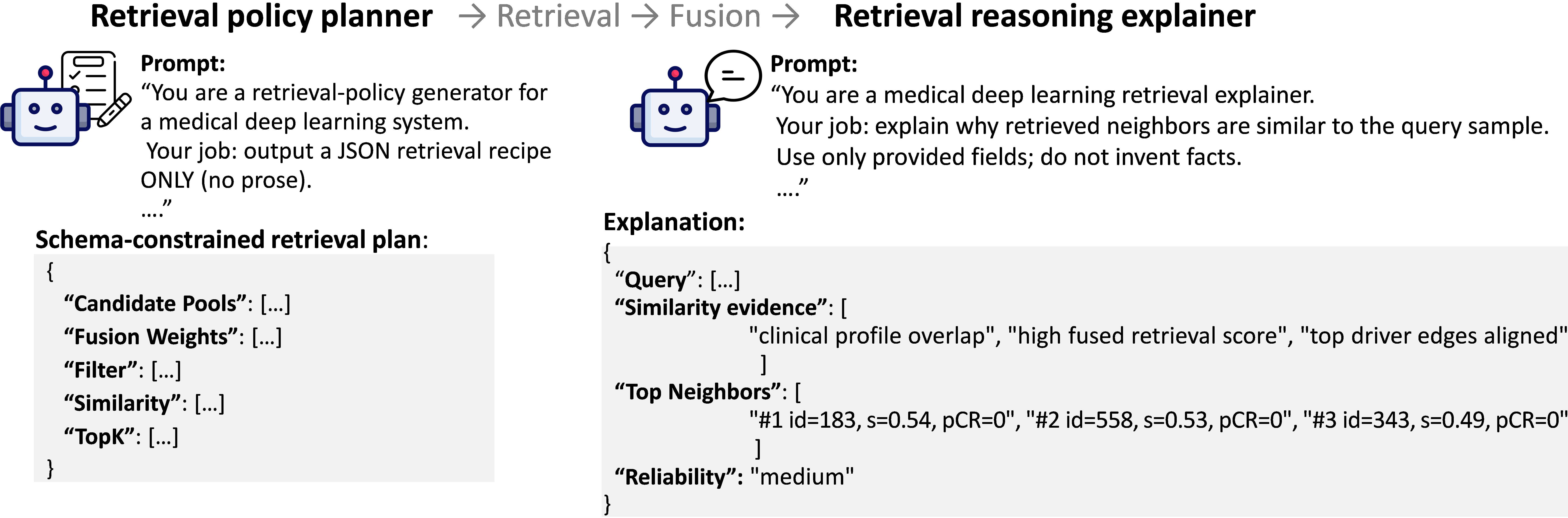}
\caption{LLM-driven RAG case study for a pCR=0 patient from external set.
} 
\label{case}
\end{figure}

\textbf{Qualitative analysis.}
Fig. \ref{SHAP} presents subtype-specific SHAP analysis of directed edges in ClinRAG-GRAPH, showing that the model captures clinically meaningful cross-modal dependencies rather than relying on isolated features. 
The influential edges vary across Luminal A/B, HER2-pos, and TripleNeg subgroups, suggesting subtype-aware message passing. 
Consistent contributions from edges linked to ER, PR, HER2, Ki67, age, and MRI features support the effectiveness of our proposed hierarchical clinical-prior graph. The directional importance patterns further indicate that the model benefits from structured relational reasoning instead of simple feature concatenation.

\textbf{LLM-driven RAG case study.}
As illustrated in Fig. \ref{case}, the Deepseek serves as both a retrieval planner and a post-hoc explainer. The planner generates a schema-constrained retrieval policy that selects clinically matched neighbors for the pCR=0 query case. The retrieved samples show high fused similarity and consistent clinical profiles. After fusion, the top-$3$ neighbors provide coherent supporting evidence with aligned driver-edge patterns. The LLM explainer then summarizes this evidence into a structured, verifiable rationale. The final prediction agrees with the retrieved evidence and remains stable for this case.

\section{Conclusion}
In this study, we explore pre-treatment pCR prediction under multimodal heterogeneity and multicenter domain shift. We propose ClinRAG-GRAPH, a clinically informed framework that integrates DCE-MRI, clinical variables, and pathological biomarkers through hierarchical graph modeling, adversarial learning, and RAG inference. Extensive experiments demonstrate improved predictive performance, cross-center generalization, and clinically interpretability. Future work will collect longitudinal data to further improve pCR prediction.

\begin{credits}
\subsubsection{\ackname} This work was supported by Shenzhen Medical Research Fund (D2501013), Macao Polytechnic University Grant (RP/FCA-13/2026), and the Science and Technology Development Fund, Macau SAR (File no. 0004/2025/ASJ) under the FDCT-FAPESP Joint Funding Scheme.
\subsubsection{\discintname}
The authors have no competing interests to declare that are relevant to the content of this article.
\end{credits}

%
\bibliographystyle{splncs04}
\bibliography{Paper-1181}

\end{document}